\pdfoutput=1

\documentclass[11pt]{article}

\usepackage{emnlp2021}
\usepackage{multirow}
\usepackage{times}
\usepackage{latexsym}
\usepackage{hyperref}
\usepackage[T1]{fontenc}

\usepackage[utf8]{inputenc}

\usepackage{microtype}

\usepackage{amsmath}
\usepackage{hyperref}
\usepackage{tikz}
\usepackage{pgfplots}
\pgfplotsset{compat=1.16}
\usepackage{multirow}
\usepackage{multicol}
\usepackage{caption}
\usepackage{graphicx}
\usepackage{tabu}
\usepackage{makecell}
\usepackage{soul}
\usepackage{booktabs}
\usepackage{array}
\usepackage{xcolor}
\usepackage{ragged2e}
\usepackage{subcaption}
\usepackage{footmisc}
\usepackage{algorithm}
\usepackage[noend]{algpseudocode}
\usepackage{xpatch}
\title{Adversarial Examples for Evaluating Math Word Problem Solvers}

\author{Vivek Kumar\thanks{\; Equal Contribution} , Rishabh Maheshwary\footnotemark[1] \ and Vikram Pudi\\
Data Sciences and Analytics Center, Kohli Center on Intelligent Systems \\
International Institute of Information Technology, Hyderabad, India \\ {\{\text{vivek.k, rishabh.maheshwary\}@research.iiit.ac.in, vikram@iiit.ac.in}}}

\begin{document}
\maketitle
\begin{abstract}
Standard accuracy metrics have shown that Math Word Problem (MWP) solvers have achieved high performance on benchmark datasets. However, the extent to which existing MWP solvers truly understand language and its relation with numbers is still unclear.  In this paper, we generate adversarial attacks to evaluate the robustness of state-of-the-art MWP solvers. We propose two methods \emph{Question Reordering} and \emph{Sentence Paraphrasing} to generate adversarial attacks. We conduct experiments across three neural MWP solvers over two benchmark datasets. On average, our attack method is able to reduce the accuracy of MWP solvers by over $40$ percentage points on these datasets. Our results demonstrate that existing MWP solvers are sensitive to linguistic variations in the problem text. We verify the validity and quality of generated adversarial examples through human evaluation.
\end{abstract}

\section{Introduction}

A Math Word Problem (MWP) consists of a natural language text which describes a world state involving some known and unknown quantities. The task is to parse the text and generate equations that can help find the value of unknown quantities. Solving MWP's is challenging because apart from understanding the text, the model needs to identify the variables involved, understand the sequence of events, and associate the numerical quantities with their entities to generate mathematical equations. An example of a simple MWP is shown in Table \ref{table:1}.
\begin{table}[h!]
\centering
\resizebox{0.45\textwidth}{!}{%
{\renewcommand{\arraystretch}{1.1}
\begin{tabular}{@{}l@{}}
\toprule
\begin{tabular}[c]{@{}l@{}}\textbf{Original Problem}\\ Text: \textcolor{red}{Tim} has 5 \textcolor{blue}{books}. \textcolor{red}{Mike} has 7 \textcolor{blue}{books}. \\ \textcolor{black}{How many books do they have together?}\\ Equation: X = 5+7\end{tabular} \\ \midrule
\begin{tabular}[c]{@{}l@{}}\textbf{Question Reordering}\\ Text: \textcolor{black}{How many books do they have together} \\  \textcolor{black}{given that Tim has 5 books and} \textcolor{black}{Mike has 7 books.}\\ Equation: X = 5*7\end{tabular} \\ \midrule
\begin{tabular}[c]{@{}l@{}}\textbf{Sentence Paraphrasing}\\ Text: \textcolor{red}{Tim} \textcolor{black}{has got 5} \textcolor{blue}{books}.  \textcolor{black}{There are 7} \textcolor{blue}{books} \textcolor{black}{in} \\ \textcolor{red}{Mike's}  \textcolor{black}{possession.} \textcolor{black}{How many books do they have?}\\ Equation: X = 5*5\end{tabular} \\ \bottomrule
\end{tabular}}}
\caption{A MWP and generated adversarial examples by our methods. Red and blue color denote the subject and the entity respectively of numerical values.}
\label{table:1}
\end{table}
In recent years, solving MWPs has become a problem of central attraction in the NLP community. There are a wide variety of MWPs ranging from simple linear equations in one variable~\cite{koncel2016mawps,miao2020diverse} to complex problems that require solving a system of equations ~\cite{huang2016well,saxton2019analysing}. In this paper, we consider simple MWPs which can be solved by a linear equation in one variable.

Existing MWP solvers can be categorized into statistical learning based ~\cite{hosseini-etal-2014-learning,kushman-etal-2014-learning} and deep learning based solvers
. However, recent deep learning based approaches  ~\cite{wang2017deep,xie2019goal,zhang2020graph} have established their superiority over statistical learning based solvers. Here, we will briefly review some recent MWP solvers.
Initially, ~\cite{wang2017deep} modelled the task of MWP as a sequence to sequence task and utilized Recurrent Neural Nets (RNNs) to learn problem representations. Building upon this, ~\citep{chiang2018semantically} focused on learning representations for mathematical operators and numbers,~\citep{xie2019goal,Wang2019TemplateBasedMW} utilized tree structure to develop decoders for MWP solvers. More recently, to learn accurate relationship between numerical quantities and their attributes ~\cite{zhang2020graph} modelled encoder as a graph structure.

All such MWP solvers have achieved high performance on benchmark datasets. However, the extent to which these solvers truly understand language and numbers remains unclear.
Prior works on various NLP tasks have shown that Deep Neural Networks (DNNs) attend to superficial cues to achieve high performance on benchmark datasets. Recently, \cite{patel-etal-2021-nlp} proposed a challenge test set called SVAMP which demonstrate that existing MWP solvers rely on shallow heuristics to achieve high performance. Instead of relying on standard accuracy metrics, many works have used adversarial examples~\cite{szegedy2013intriguing,papernot2017practical} to evaluate the robustness of neural NLP models.
Adversarial examples are generated by making small changes to the original input such that the adversarial example is $(1)$ semantically similar to the original input, $(2)$ is grammatically correct and fluent and $(3)$ deceives the DNNs to generate an incorrect prediction.

In~\cite{jia2017adversarial} authors crafted adversarial attacks to test the robustness of QA systems. Prior works in ~\cite{glockner2018breaking,mccoy2019right} uses adversarial examples to show deficiencies of NLI models. Similarly,~\cite{dinan2019build,cheng2019robust} uses adversarial examples to develop robust dialogue and neural machine translation models.  Recently, there has been a plethora of work~\cite{ebrahimi2017hotflip,alzantot2018generating,jin2020bert,maheshwary2021generating,maheshwary2020context} to evaluate text classification systems against adversarial examples. Although adversarial examples are commonly used for various NLP tasks, there has been no work that uses adversarial examples to evaluate MWP solvers.
In this paper, we bridge this gap and evaluate the robustness of state-of-the-art MWP solvers against adversarial examples.

Generating adversarial attacks for MWP is a challenging task as apart from preserving textual semantics, numerical value also needs to be preserved. The text should make mathematical sense, and the sequence of events must be maintained such that humans generate the same equations from the problem text. Standard adversarial generation techniques like synonym replacement~\citep{alzantot2018generating} are not suitable for MWP as the fluency of the problem statement is not preserved. Similarly, paraphrasing techniques like back-translation ~\citep{mallinson-etal-2017-paraphrasing} are not ideal as they generate syntactically uncontrolled examples.

We propose two methods to generate adversarial examples on MWP solvers, $(1)$ Question Reordering --- It transforms the problem text by moving the question part to the beginning of the problem and $(2)$ Sentence Paraphrasing --- It paraphrases each sentence in the problem such that the semantic meaning and the numeric information remains unchanged. Our results demonstrate that current solvers are not robust against adversarial examples as they are sensitive to minor variations in the input. We hope that our insights will inspire future work to develop more robust MWP solvers.
Our contributions are as follows:
\begin{enumerate}

\item To the best of our knowledge, this is the first work that evaluates the robustness of MWP solvers against adversarial attacks. We propose two methods to generate adversarial examples on three MWP solvers across two benchmark datasets.

\item On average, the generated adversarial examples are able to reduce the accuracy of MWP solvers by over $40\%$. Further, we experiment with different type of input embeddings and perform adversarial training using our proposed methods. We also conducted human evaluation to ensure that the generated adversarial examples\footnote{Adversarial Examples and Code is available at: \href{https://github.com/kevivk/MWP_Adversarial}{https://github.com/kevivk/MWP\_Adversarial}} are valid, semantically similar and grammatically correct.
\end{enumerate}
\section{Proposed Approach}
\subsection{Problem Definition}
A MWP is defined as an input of $n$ tokens, $\mathcal{P} = \{w_1,w_2..w_n\}$ where each token $w_i$ is either a numeric value or a word from a natural language. The goal is to generate a valid mathematical equation $\mathcal{E}$ from $\mathcal{P}$ such that the equation consists of numbers from $\mathcal{P}$, desired numerical constants and mathematical operators from the set $\{/,*,+,-\}$. The above problem can also be expressed as $\mathcal{P} = \{S_1,S_2..S_k,Q\}$ where $Q$ is the question, $\{S_1,S_2..S_k\}$ are the sentences constituting the problem description.\\
Let $\textbf{F}: \mathcal{P} \rightarrow \mathcal{E}$ be a MWP solver where $\mathcal{E}$ is the solution equation to problem $\mathcal{P}$. Our goal is to craft an adversarial text input $\mathcal{P}^*$ from the original input $\mathcal{P}$ such that the generated sequence is $(1)$ semantically similar to the original input, $(2)$ preserves sequence of events in the problem, $(3)$ preserve numerical values and $(4)$ makes the MWP solver $\textbf{F}$ to generate an incorrect equation $\mathcal{E}^*$ for the unknown variable.
We assume a black-box setting in which we have no access to the parameters, architecture or training data of the MWP solver. We only have access to the input text and equations generated by the solver.

\subsection{Question Reordering}
To examine whether existing MWP solvers are sensitive to the order of the question in the problem text, we moved the question $Q$ at the start, followed by the rest of the problem description $\{S_1,S_2..S_k\}$. Formally, given the original input $\mathcal{P} = \{S_1,S_2...S_k,Q\}$ we transformed this to $\mathcal{P}^* = \{Q,S_1,S_2...S_k\}$. We keep the rest of the problem description $\{S_1,S_2..S_k\}$ unaltered. Also, to ensure that the generated problem text $\mathcal{P}^*$ is grammatically correct and fluent, we added phrases like "Given that" or "If" after the end of the question $Q$ and before the start of the sentences $\{S_1,S_2..S_k\}$. An example of this is shown in Table $1$. We additionally, make use of co-reference resolution and named entity recognition\footnote {\href{https://spacy.io/}{https://spacy.io/}} to replace pronouns with their co-referent links. Note that placing the question $Q$ at the start rather than any other position ensures that the generated problem $\mathcal{P}^*$ has the same sequence of events as the original problem $\mathcal{P}$. Moreover, this method is better than randomly shuffling the sentences in $\mathcal{P}$ as it can change the sequence of events in the problem, resulting in a completely different equation.

\subsection{Sentence Paraphrasing}
To check whether MWP solvers generate different equations to semantically similar inputs, we generate paraphrases of each sentence in the problem text. Sentence Paraphrasing ensures that solvers do not generate equations based on keywords and specific patterns. Formally, given a problem statement $\mathcal{P}$ we obtain top $m$ paraphrases for each sentence $S_i$ as $\{S_{i,1},S_{i,2},...,S_{i,m}\}$ and for question $Q$  as $\{Q_{i,1},Q_{i,2},...,Q_{i,m}\}$ by passing it through a paraphrasing model $\mathcal{M}$. For sentences with numerical values present in them, we need to ensure that each paraphrase candidate associates the numeric values with the same entity and subject as it is present in the original sentence $S_i$. To ensure this, we follow the approach used in ~\citep{hosseini-etal-2014-learning} to segregate each sentence $S_i$ into entities and its subject. These are collectively labeled as head entity $h_{i,orig}$ for the original sentence $S_i$ and $h_{i,k}$ for the paraphrase candidates $S_{i,k}$. This methodology ensures that each numeric value is still associated correctly with its attributes even after paraphrasing. Paraphrased sentences that do not have matching head entities for any of the numeric values are filtered out.
The remaining paraphrases of $S_i$ and question $Q$ are combined to generate all possible combinations of problem texts. The input combination for which the MWP solver generates an incorrect or invalid equation is selected as the final adversarial problem text $\mathcal{P}^*$. Sentence Paraphrasing generates inputs containing small linguistic variations and diverse keywords (more examples in appendix). Therefore, it is used to evaluate whether existing MWP solvers rely on specific keywords or patterns to generate equations. Algorithm $1$ shows all the steps followed above to generate paraphrases.
\begin{algorithm}[h!]
\caption{Sentence Paraphrasing}
\textbf{Input: }Problem text $\mathcal{P} ,\mathcal{M}$ is Paraphrase model\\
\textbf{Output: }Adversarial text $\mathcal{P}^*$
\begin{algorithmic}[1]
\State $\mathcal{P}^* \gets \mathcal{P}$
\State $y_{orig} \gets \bf{F}(\mathcal{P})$
\For {$S_i\ in\ \mathcal{P}$}
\State ${C} \gets\ \mathcal{M}(S_{i})$
\For {$c_{j}\ in\ {C}$}
\If{$h_{i,orig} == h_{i,j}$}
\State $paraphrases.add(c_j)$\
\EndIf
\EndFor
\State $paraphrases.add(S_i)$
\State $candidates.add(paraphrases)$
\EndFor
\For{$c_k$\ in \ $Combinations(candidates)$}
\State $y_{adv}\ \gets \ $\bf{F}$(c_k)$
\If {$y_{adv}\ \neq y_{orig}$}
\State $\mathcal{P}^* \gets c_k $
\State $\textbf{end}$
\EndIf
\EndFor
\end{algorithmic}
\label{algorithm:1}
\end{algorithm}

\section{Experiments}
\subsection{Datasets and Models}
We evaluate the robustness of three state-of-the-art MWP solvers: $(1)$ \emph{Seq2Seq}~\cite{wang2017deep} having an LSTM encoder and an attention based decoder. $(2)$ \emph{GTS}~\cite{xie2019goal} having an LSTM encoder and a tree based decoder and $(3)$ \emph{Graph2tree}~\cite{zhang2020graph} consists of a both a tree based encoder and decoder. Many existing datasets are not suitable for our analysis as either they are in Chinese~\citep{wang2017deep} or they have problems of higher complexities ~\citep{huang2016well} . We conduct experiments across the two largest available English language datasets satisfying our requirements: $(1)$ \emph{MaWPS}~\cite{koncel2016mawps} containing
$2,373$ problems $(2)$ \emph{ASDIV-A}~\cite{miao2020diverse} containing $1,213$ problems. Both datasets have MWPs with linear equation in one variable.
\subsection{Experimental Setup}
We trained the three MWP solvers from scratch as implemented in baseline paper \citep{wang2017deep} on the above two datasets using 5-fold cross-validation as followed in~\cite{zhang2020graph}. The original accuracies obtained on the datasets are shown in Table $2$. We used~\cite{zhang2020pegasus} to generate paraphrases of each sentence in the problem text. Same hyperparameter values were used as present in the original implementation of the paraphrase model. We conducted a human evaluation (Section $4.3$) to verify the quality of generated adversarial examples. Further details are given in Appendix.
\subsection{Implementation Details} For conducting our experiments we have used two Boston SYS-7048GR-TR nodes equipped with NVIDIA GeForce GTX 1080 Ti computational GPU’s . The number of parameters ranged from 20M to 130M for different models. Hyper-parameter values were not modified, and we follow the recommendations of the respective models.
We chose the number of candidate paraphrases $m$ used in Algorithm \ref{algorithm:1} to be 7. Generating adversarial examples using Question Reordering took around 3 minutes on average for both MaWPS and ASDiv-A dataset. Sentence Paraphrasing took around 10 minutes on average for generation of adversarial examples on both the datasets. These experiments are not computation heavy as the generation technique is of linear order and number of
examples are moderate.
\subsection{Results}
Table $2$ shows the results of our proposed methods. On average, the generated adversarial examples can lower the accuracy of MWP solvers by over $40$ percentage points. Across both datasets, \emph{Graph2Tree}, the state-of-the-art MWP solver achieves only $34\%$ and $24\%$ accuracy on \emph{Question Reordering} and \emph{Sentence Paraphrasing} respectively. \emph{Sentence Paraphrasing} is around $10$ percentage points more successful in attacking MWP solvers than \emph{Question Reordering}. These results verify our claim that current MWP solvers are sensitive to small variations in the input. Table $1$ shows an MWP problem and its adversarial counterparts generated by our method—more examples in the Appendix section.
\begin{table}[]
\small
{\renewcommand{\arraystretch}{0.9}
\begin{tabular}{@{}c|c|c|c|c@{}}
\toprule
\textbf{Dataset} & \textbf{Eval Type} & \textbf{Seq2Seq} & \textbf{GTS} & \textbf{Graph2Tree} \\ \midrule
\multirow{3}{*}{\textbf{MaWPS}} & Orig & 53.0 & 82.6 & 83.7 \\ \cmidrule(l){2-5} 
 & QR & \bf{18.2} & \bf{32.3} & \bf{35.6} \\ \cmidrule(l){2-5} 
 & SP & \textbf{10.5} & \textbf{22.7} & \textbf{25.5} \\ \midrule
\multirow{3}{*}{\textbf{ASDIV-A}} & Orig & 54.5 & 71.4 & 77.4 \\ \cmidrule(l){2-5} 
 & QR & \bf{17.5} & \bf{30.5} & \bf{33.5} \\ \cmidrule(l){2-5} 
 & SP & \textbf{13.2} & \textbf{21.2} & \textbf{23.8} \\ \bottomrule
\end{tabular}}
\caption{Results of MWP Solvers on adversarial examples. Orig is the original accuracy, QR is  Question Reordering and SP is Sentence Paraphrasing.}
\label{table:2}
\end{table}
\section{Analysis}

\subsection{BERT Embeddings} We trained the solvers using pre-trained BERT embeddings and then generated adversarial examples against them using our proposed methods. Results obtained are shown in Table $3$. We see that using BERT embeddings, the original accuracy of MWP solvers increases by $5$ percentage points, and they are more robust than solvers trained from scratch. Specifically, these solvers do well against \emph{Question Reordering} because of the contextualized nature of BERT embeddings, but for examples generated using \emph{Sentence Paraphrasing} methods these models do not perform well. However, on average, our adversarial examples can lower the accuracy by $30$ percentage points on both datasets.
\subsection{Adversarial Training} To examine the robustness of MWP solvers against our attacks, we generated adversarial examples on the training set of both the datasets using our proposed methods and then augmented the training sets with the generated adversarial examples. We then retrained the MWP solvers and again attacked these solvers using our methods. Table $3$ shows that the MWP solvers become more robust to attacks. Specifically, the solvers perform well against \emph{Question Reordering} but are still deceived by \emph{Sentence Paraphrasing}. Nevertheless, our proposed attack methods are still able to lower the accuracy of MWP solvers by $25$ percentage points.
\subsection{Human Evaluation}
To verify the quality and the validity of the adversarial examples, we asked human evaluators $(1)$ To check if the paraphrases will result in the same linear equation as that of the original problem, $(2)$ Evaluate each adversarial example in the range $0$ to $1$ to check its semantic similarity with the original problem and $(3)$ On a scale of $1$ to $5$ rate each adversarial example for its grammatical correctness. We also explicitly check for examples which do not satisfy our evaluation criteria and manually remove them from adversarial examples set. Three different human evaluators evaluate each sample, and the mean results are shown in Table $4$. 
\begin{table}[]
\small
{\renewcommand{\arraystretch}{0.9}
\begin{tabular}{@{}c|c|c|c|c@{}}
\toprule
\textbf{Dataset} & \textbf{Eval Type} & \textbf{Seq2Seq} & \textbf{GTS} & \textbf{Graph2Tree} \\ \midrule
\multirow{4}{*}{\textbf{MaWPS}} 
 & Adv (QR) & {32.4} & {52.3} & {54.9} \\ \cmidrule(l){2-5} 
  & Adv (SP) & {27.6} & {40.7} & {42.3} \\ \cmidrule(l){2-5}
  & BERT (QR) & \bf{45.3} & \bf{63.0} & \bf{65.6} \\ \cmidrule(l){2-5}
 & BERT (SP) & \textbf{32.5} & \textbf{43.5} & \textbf{45.5} \\
 \midrule
\multirow{4}{*}{\textbf{ASDIV-A}}
 & Adv (QR) & {34.5} & {48.4} & {54.8} \\ \cmidrule(l){2-5} 
 & Adv (SP) & {28.8} & {31.6} & {33.0} \\ \cmidrule(l){2-5}
 & BERT (QR) & \bf{41.3} & \bf{59.8} & \bf{62.7} \\ \cmidrule(l){2-5}
 & BERT (SP) & \textbf{30.6} & \textbf{40.0} & \textbf{42.6} \\ \bottomrule
\end{tabular}}
\caption{Accuracy of MWP solvers with adversarial training on our proposed methods. Adv and BERT represent models trained from scratch and BERT embeddings respectively.}
\label{table:3}
\end{table}

\begin{table}[h!]
\small
\centering
{\renewcommand{\arraystretch}{0.5}
\begin{tabular}{@{}c|c|c@{}}
\toprule
\textbf{Evaluation criteria} & \textbf{MaWPS} & \textbf{ASDIV-A} \\ \midrule
Same Linear Equation & 85.7\% & 86.2\% \\ \cmidrule(l){1-3} 
Semantic Similarity & 0.88 & 0.89 \\ \cmidrule(l){1-3} 
Grammatical Correctness & 4.55 & 4.63 \\ \bottomrule
\end{tabular}}
\caption{Human Evaluation scores on datasets}
\label{table4}
\end{table}

\section{Future Work and Conclusion}
The experiments in this paper showcase that NLP models do not understand MWP entirely and are not robust enough for practical purposes. Our work encourages the development of robust MWP solvers and techniques to generate adversarial math examples. We believe that the generation of quality MWP's will immensely help develop solvers that genuinely understand numbers and text in combination. Future works could focus on creating more such techniques for adversarial examples generation and making robust MWP solvers.

\section{Acknowledgments}
We would like to thank the anonymous reviewers for their constructive feedback. We would also like to thank our colleagues at Data Sciences and Analytics Center, IIIT Hyderabad for providing valuable feedback. Special thanks to the human annotators who have helped in evaluation of generated adversarial examples.

\bibliography{anthology,custom}
\bibliographystyle{acl_natbib}
\clearpage
\appendix

\begin{table*}[]
\centering
\resizebox{\textwidth}{!}{%
{\renewcommand{\arraystretch}{0.9}
\begin{tabular}{@{}l@{}}
\toprule
\begin{tabular}[c]{@{}l@{}}\textbf{Original Problem}\\ \textcolor{red}{Problem Statement :} A teacher had 7 worksheets to grade . If she graded 3 , but then another 4 were turned in,\\ how many
worksheets would she have to grade ? \\
\textcolor{blue}{Predicted Equation :} X = 7+3-4\\
\textbf{Question Reordering}\\
\textcolor{red}{Problem Statement :} How many worksheets would she have to grade given that a teacher had 7 worksheets\\ to grade and if she
graded 3 but then another 4 were turned in?\\
\textcolor{blue}{Predicted Equation :} X = 7+3+4\\
\textbf{Sentence Paraphrasing}\\
\textcolor{red}{Problem Statement :} A teacher had her students work on 7 questions. 3 would be graded if she graded it.\\ Then another 4 was
turned in. How many things would she have to grade?\\
\textcolor{blue}{Predicted Equation :} X = 7+3-4
\end{tabular} \\ \midrule 
\begin{tabular}[c]{@{}l@{}}\textbf{Original Problem}\\ \textcolor{red}{Problem Statement :} Gwen earned 20 points for each bag of cans she recycled . If she had 10 bags, but didn’t\\ recycle 3 of them , how many points would she have earned ? \\
\textcolor{blue}{Predicted Equation :} X = (20*(10-3)) \\
\textbf{Question Reordering}\\
\textcolor{red}{Problem Statement :} How many points would she have earned given that Gwen earned 20 points for each bag\\of cans she recycled and if she had 10 bags but didn’t recycle 3 of them ?\\
\textcolor{blue}{Predicted Equation :} X = 20*10-3 \\
\textbf{Sentence Paraphrasing}\\
\textcolor{red}{Problem Statement :} Gwen earned 20 points for each bag of cans she recycled. She have 10 bags.\\ She did not recycle 3 of them. How many points would she have gotten?\\
\textcolor{blue}{Predicted Equation :} X = 20+10-3
\end{tabular} \\ \midrule 
\begin{tabular}[c]{@{}l@{}}\textbf{Original Problem}\\ \textcolor{red}{Problem Statement :} : Dennis has 12 pencils stored in boxes. If there are 3 boxes, how many pencils must\\ go in each box?\\
\textcolor{blue}{Predicted Equation :} X = 12/3 \\
\textbf{Question Reordering}\\
\textcolor{red}{Problem Statement :} : If there are 3 boxes, how many pencils must go in each box given that Dennis has\\ 12 pencils stored in boxes ?\\
\textcolor{blue}{Predicted Equation :} X = 12/3\\
\textbf{Sentence Paraphrasing}\\
\textcolor{red}{Problem Statement :}  Dennis has 12 pencils in boxes. There are 3 boxes. Find the number of pencils in each box?\\
\textcolor{blue}{Predicted Equation :} X = 12-3
\end{tabular} \\ \midrule
\begin{tabular}[c]{@{}l@{}}\textbf{Original Problem}\\ \textcolor{red}{Problem Statement :} Oliver made 10 dollars mowing lawns over the summer . If he spent 4 dollars buying new\\ mower blades. How many 3 dollar games could he buy with the money he had left ?\\
\textcolor{blue}{Predicted Equation :} X = (10-4)/3  \\
\textbf{Question Reordering}\\
\textcolor{red}{Problem Statement :} How many 3 dollar games could Oliver buy with the money he had left given that Oliver\\ made 10 dollars mowing lawns over the summer and if he spent 4 dollars buying new mower blades.\\
\textcolor{blue}{Predicted Equation :} X = (10-4)*3\\
\textbf{Sentence Paraphrasing}\\
\textcolor{red}{Problem Statement :} Over the summer, Oliver made 10 dollars mowing lawns. He spent 4 dollars on new blades.\\ With the money he had left, how many 3 dollar games could he buy?\\
\textcolor{blue}{Predicted Equation :} X = (10-4)*3
\end{tabular} \\ \midrule
\end{tabular}}}
\caption{Some instances of valid Adversarial Examples}
\label{tabel:example}
\end{table*}

\begin{table*}[t]
\centering
\resizebox{\textwidth}{!}{%
{\renewcommand{\arraystretch}{0.9}
\begin{tabular}{@{}l@{}}
\toprule
\begin{tabular}[c]{@{}l@{}}\textbf{Original Problem}\\ \textcolor{red}{Problem Statement :}  A trivia team had 10 members total. But during a game 2 members did not show up. If each\\ member that did show up scored 3 points. How many points were scored?\\
\textcolor{blue}{Predicted Equation :} X = (10-2)*3\\
\textbf{Sentence Paraphrasing}\\
\textcolor{red}{Problem Statement :} A team with 10 members had a lot of questions to answer. But during the game 2 members\\ did not show up. 3 points were scored if each member showed up. How many points were scored?\\
\end{tabular} \\ \midrule 
\begin{tabular}[c]{@{}l@{}}\textbf{Original Problem}\\ \textcolor{red}{Problem Statement :} A tailor cut 15 of an inch off a skirt and 5 of an inch off a pair of pants . How much more did\\ the tailor cut off the skirt than the pants ?\\
\textcolor{blue}{Predicted Equation :} X = 15-5\\
\textbf{Sentence Paraphrasing}\\
\textcolor{red}{Problem Statement :} The 15 was cut by a tailor. There is a skirt and 5 of an inch off. There is a pair of pants.\\ How much more did the tailor cut off the skirt than the pants?\\
\end{tabular} \\ \midrule
\begin{tabular}[c]{@{}l@{}}\textbf{Original Problem}\\ \textcolor{red}{Problem Statement :} A vase can hold 10 flowers . If you had 5 carnations and 5 roses,\\ how many vases would you need to hold the flowers?\\
\textcolor{blue}{Predicted Equation :} X = (5+5)/10\\
\textbf{Sentence Paraphrasing}\\
\textcolor{red}{Problem Statement :} 10 flowers can be held in a vase. If you had 5 and 5 roses. How many vases\\ do you need to hold the flowers.
\end{tabular} \\ \midrule
\end{tabular}}}
\caption{Some instances of invalid Adversarial Examples}
\label{tabel:invalid}
\end{table*}

\end{document}